\documentclass[10pt,twocolumn,letterpaper]{article}
\usepackage{cvpr}      

\usepackage{graphicx}
\usepackage{amsmath}
\usepackage{amssymb}
\usepackage{booktabs}
\usepackage{multirow}
\usepackage{bm}
\usepackage{paralist} 
\usepackage[accsupp]{axessibility} 

\usepackage[pagebackref,breaklinks,colorlinks]{hyperref}

\usepackage[capitalize]{cleveref}
\crefname{section}{Sec.}{Secs.}
\Crefname{section}{Section}{Sections}
\Crefname{table}{Table}{Tables}
\crefname{table}{Tab.}{Tabs.}

\def\cvprPaperID{9018} 
\def\confName{CVPR}
\def\confYear{2022}
\begin{document}

\title{Learning Non-target Knowledge for Few-shot Semantic Segmentation}

\author{
 Yuanwei Liu$^{1}$
 \hspace{15pt}
 Nian Liu$^{2}$\footnote[1]{}
 \hspace{15pt}
 Qinglong Cao$^{1}$ 
 \hspace{15pt}
 Xiwen Yao$^{1}$
 \hspace{15pt}
 Junwei Han$^{1}$ 
 \hspace{15pt}
 Ling Shao$^{3}$
 \hspace{15pt}
 \\
 $^1$Northwestern Polytechnical University
  \hspace{15pt}
 $^2$Inception Institute of Artificial Intelligence
 \\
 $^3$Terminus Group, China
 \\
 {\tt\small
    \{liuyuanwei9809, liunian228, caoql19980603, yaoxiwen517, junweihan2010\}@gmail.com
    }\\
 {\tt\small
 ling.shao@ieee.org
    }
}

\maketitle

\footnotetext[1]{Corresponding author.} 


\begin{abstract}
	Existing studies in few-shot semantic segmentation only focus on mining the target object information, however, often are hard to tell ambiguous regions, especially in non-target regions, which include background (BG) and Distracting Objects (DOs). To alleviate this problem, we propose a novel framework, namely Non-Target Region Eliminating (NTRE) network, to explicitly mine and eliminate BG and DO regions in the query. First, a BG Mining Module (BGMM) is proposed to extract the BG region via learning a general BG prototype. To this end, we design a BG loss to supervise the learning of BGMM only using the known target object segmentation ground truth. Then, a BG Eliminating Module and a DO Eliminating Module are proposed to successively filter out the BG and DO information from the query feature, based on which we can obtain a BG and DO-free target object segmentation result. Furthermore, we propose a prototypical contrastive learning algorithm to improve the model ability of distinguishing the target object from DOs. Extensive experiments on both PASCAL-${5^{i}} $ and COCO-$ 20^{i} $ datasets show that our approach is effective despite its simplicity. Code is available at \href{https://github.com/LIUYUANWEI98/NERTNet}{https://github.com/LIUYUANWEI98/NERTNet}
\end{abstract}

\vspace{-3mm}
\section{Introduction}

Due to the rapid development of fully convolutional network (FCN) \cite{long2015fully} architectures, deep learning has made milestone progress in semantic segmentation. Most methods adopt the fully-supervised learning scheme and require a mass of annotated data for training. Although they can achieve good performance, their data-hungry nature demands time and labor-consuming image annotations. To alleviate this problem, few-shot semantic segmentation was proposed to segment unseen object classes in query images with only a few annotated samples, namely supports.

\begin{figure}[htbp]
	\centering
	\includegraphics[scale =0.68]{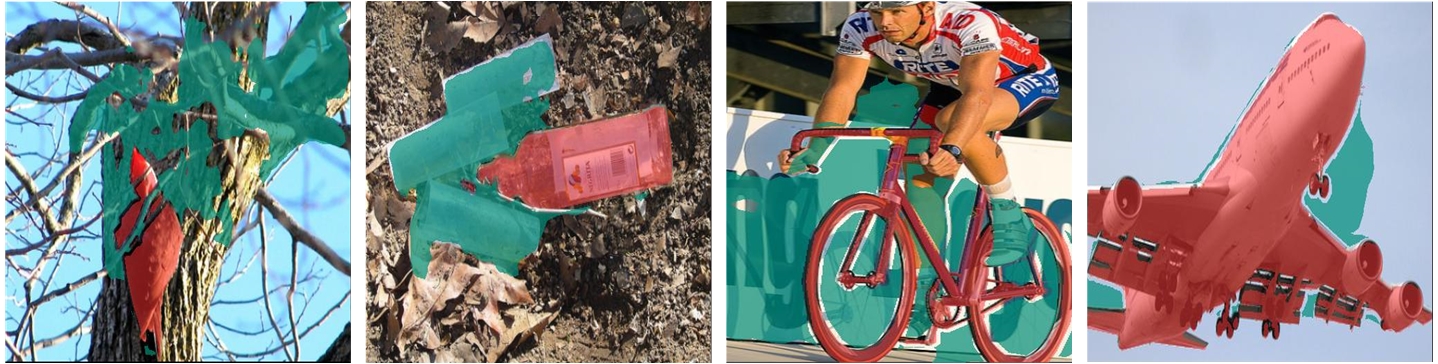}
	\vspace{-3mm}
	\caption{Previous methods often show false positive predictions in non-target regions. 
		Pixels in red indicate the target objects, while pixels in green mean false positive predictions.}
	\label{figure1}
\end{figure}

\begin{figure}[htbp]
	\centering
	\includegraphics[scale= 0.32]{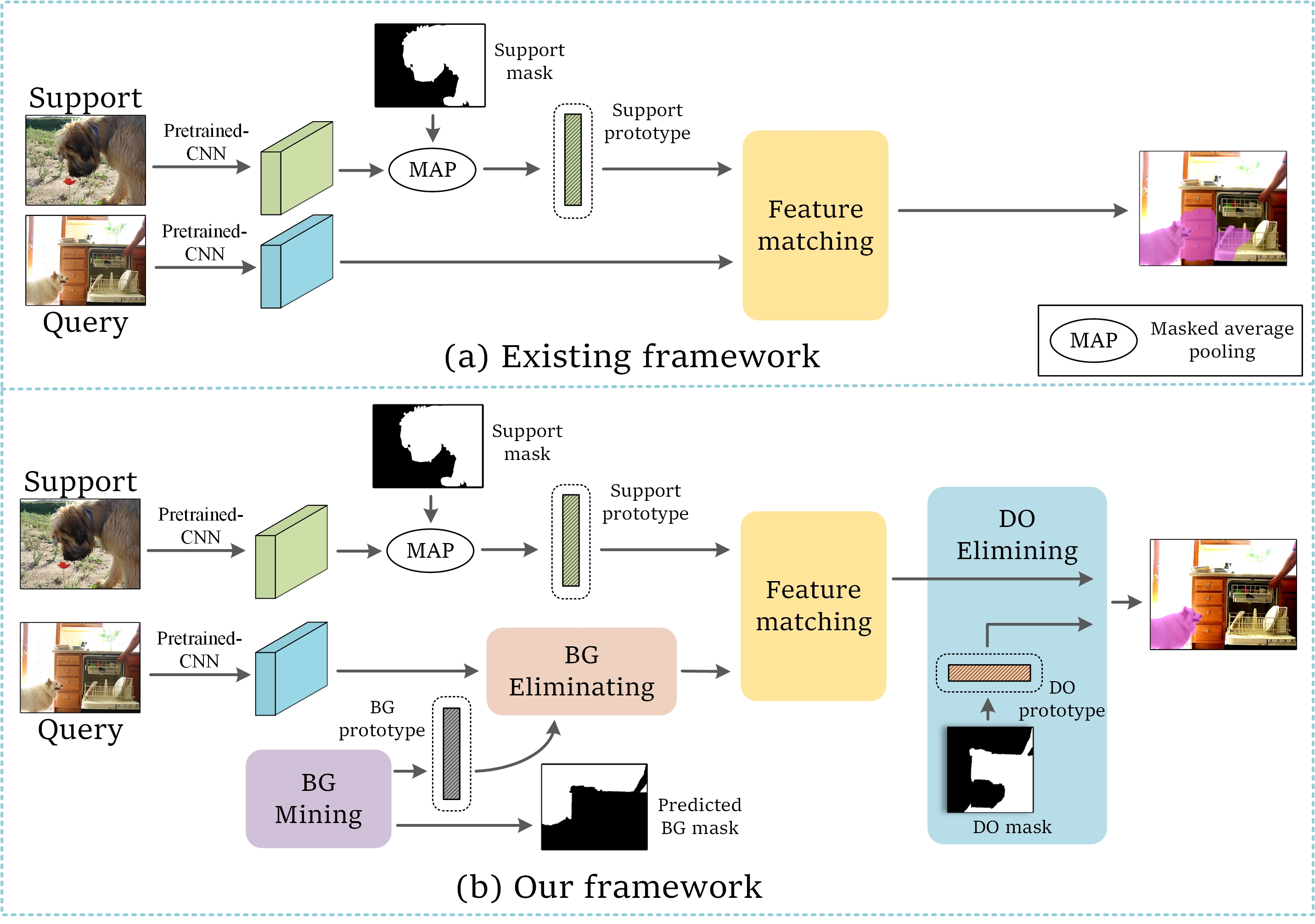}
	\vspace{-3mm}
	\caption{Comparison between existing framework and ours for few-shot segmentation. The main difference is that the former only mines target category information, while we propose to eliminate co-existing pixels belonging to non-target regions, including the background (BG) and distracting objects (DO). }
	\label{figure2}
	\vspace{-3mm}
\end{figure}

Currently, there are many existing researches exploring various deep learning methods for few-shot semantic segmentation\cite{tian2020prior,li2021adaptive,snell2017prototypical,zhang2019canet,liu2020part,wang2019panet}. 
They usually extract features from both query and support images first, and then extract the class-specific representation using the support masks. Finally, a matching network is leveraged to segment the target object in the query image using the class representation.

Most typically, prototypical learning methods \cite{dong2018few,tian2020prior,zhang2019canet,wang2019panet,zhang2020sg} use masked average pooling (MAP) on the target object regions of the support images to form a single or a few prototypes. Then, prototypes are used to segment the target object in the query image via conducting dense feature matching.

Although some achievements have been made, these methods all focus on digging out more effective target information from supports as much as possible, and then transferring them to the query image to achieve segmentation (see Figure~\ref{figure2} (a)). However, as illustrated in Figure~\ref{figure1}, they often suffer from the false positive prediction in backgrounds (BG) and co-existing objects belonging to other classes, namely, distracting objects (DOs). The main reason is that solely focusing on target objects in the few-shot setting makes their models hard on learning discriminative features and differentiating ambiguous regions.

To alleviate this problem, we rethink the few shot semantic segmentation task from a new perspective, that is, mining and excluding non-target regions, \ie, BG and DO regions, rather than directly segmenting the target object. From this point, in this paper, we propose a novel framework, namely non-target region eliminating (NTRE) network for few-shot semantic segmentation. As shown in Figure~\ref{figure2} (b), we first develop a BG mining module (BGMM) to obtain a BG prototype and segment the BG region. Then, a BG eliminating module (BGEM) is proposed to filter out the BG information from the query feature. Next, the target prototype from the support is utilized in a matching network to activate the target object in the query feature.
Subsequently, we adopt a DO eliminating module (DOEM) to mine the DO region first and then filter out the DO information from the query feature. As such, finally, we can obtain an accurate target segmentation result without the distraction from the BG and DO regions.

In the BGMM, obtaining the BG prototype is not straightforward as we obtain the support prototype. Considering that BG regions universally exist in almost every image, such as sky, grass, walls, and etc, we propose to learn a general BG prototype from both query and support images in the training set. Based on this prototype, we can segment the BG regions for all images easily. Since having no ground truth BG segmentation masks to supervise the model learning, we specifically design a BG mining loss based on the known target segmentation masks.


Furthermore, considering that it's hard to learn a good prototype feature embedding space to differentiate DOs from the target object under the few-shot setting, 
we propose the prototypical contrastive learning (PCL) method to improve the object-discrimination ability of the network by refining the prototype feature embeddings. Specifically, for a query target prototype, we treat the corresponding support target prototype as the positive sample, while the DO prototypes both in query and support are considered as negative samples. We then propose a PCL loss to enforce the prototype embeddings to be similar within the target prototypes and dissimilar between target and DO prototypes. As such, the PCL could effectively help the network distinguish target objects from DOs.


In summary, our contributions are as follows:
\begin{compactitem}
\item To the best of our knowledge, this is the first time to mine and eliminate non-target regions, including BG and DOs, for few-shot semantic segmentation, which can effectively decrease false positive predictions.

\item We propose the BGMM, BGEM, and DOEM for effectively implementing the mining and eliminating of the BG and DO regions. A novel BG mining loss is also proposed for training the BGMM without using BG ground truth.

\item We propose a PCL method to improve the model ability for better distinguishing target objects from DOs.

\item Extensive experiments on PASCAL-${5^{i}} $ and COCO-$ 20^{i}  $ show that our proposed framework yields a new state-of-the-art performance, especially on the 1-shot setting.
\end{compactitem}

\vspace{-1mm}
\section{Related Works}
\vspace{-1mm}
\paragraph{Semantic Segmentation.}
Compared with convolutional neural networks (CNNs) \cite{krizhevsky2012imagenet}, the emergence of the FCN\cite{long2015fully} has brought great progress to the semantic segmentation task. Specifically, the fully connected layers in CNNs are replaced by fully convolutional layers to enable pixel-level prediction. Based on the FCN, various network architectures are proposed to tackle the semantic segmentation problem in recent works. For example, \cite{choi2020cars,fu2019dual,li2019expectation,tao2020hierarchical,yuan2018ocnet,zhang2019acfnet,zhu2019asymmetric} propose various attention mechanisms embedded in the FCN architecture.
Some other works utilize different feature fusion methods, such as the pyramid pooling module\cite{zhao2017pyramid}, dilated convolution kernels\cite{chen2018encoder}, multi-scale feature aggregation\cite{he2019adaptive}, dense atrous spatial pyramid pooling (ASPP)\cite{yang2018denseaspp}. However, these traditional semantic segmentation networks are powerless when dealing with unseen categories. Meanwhile, training such networks are computationally costly and also requires labor-consuming pixel-level annotations on large-scale data.
\vspace{-4mm}
\paragraph{Few-shot Semantic Segmentation.}
Few-shot semantic segmentation aims to segment unseen object classes in query images with only a few annotated samples. There are two mainstream frameworks to segment the target objects in query images. One is the pixel-level matching framework, which is firstly proposed by \cite{shaban2017one}. This framework usually generates the target object prototype from the support features first, and then segments the query using dense feature matching.
Another is the pixel-level measurement framework, proposed by \cite{zhang2020sg}, which measured the embedding similarity between the query and the supports. Following the first framework, CANet\cite{zhang2019canet} utilized an Iterative Optimization Module (IOM) to refine the prediction progressively after concatenating the support prototype and the query features. PFENet\cite{tian2020prior} proposed a prior mask by calculating the cosine similarity between the support and query images on high-level features without learnable parameters. ASGNet \cite{li2021adaptive} proposed a superpixel-guided clustering method to obtain multi-part prototypes from the support and used an allocation strategy to reconstruct the support feature map instead of using prototype expanding. Following the second framework, PANet \cite{wang2019panet} embedded different object classes into different prototypes with a pre-trained encoder. Then, the query image was labeled based on the distance between the representations of the query image and the prototypes. Yang \etal \cite{yang2021mining} proposed a novel joint-training framework via introducing additional base category prototypes to mine latent novel classes during training. Most of these previous methods focus on directly segmenting the target object. Differently, in this paper, we are the first to propose leveraging complementary non-target knowledge and eliminating distracting regions for few-shot segmentation.

\vspace{-4mm}
\paragraph{Contrastive Learning} Most previous computer vision researches focus on designing artificially preferred network architectures to tackle various computer vision tasks. The emergence of contrastive learning \cite{he2020momentum} brings our focus back to mining better deep feature representations via contrasting positive and negative samples. SimCLR \cite{chen2020simple} proposed a simple self-supervised contrastive learning paradigm by using different data augmentation methods to form positive and negative samples for each image instance. MoCo\cite{he2020momentum,chen2020improved} proposed to store negative samples using a dynamically updated queue, in which only the stored feature vectors from recent batches are used for training. As such, MoCo solved the inconsistency problem of the sampled features due to the optimization to the encoder. Very recently, Wang \etal \cite{wang2021exploring} introduced contrastive learning in supervised semantic segmentation and proposed a pixel-wise contrastive algorithm. They treated the pixel embeddings of the same class and different classes as positive and negative samples, respectively. Different from them, in our work, we propose the PCL scheme to improve the objects-discrimination ability of the extracted prototypes, which could effectively help the network distinguish target objects from DOs.

\begin{figure*}[htbp]
	\begin{center}
		\includegraphics[scale = 0.95]{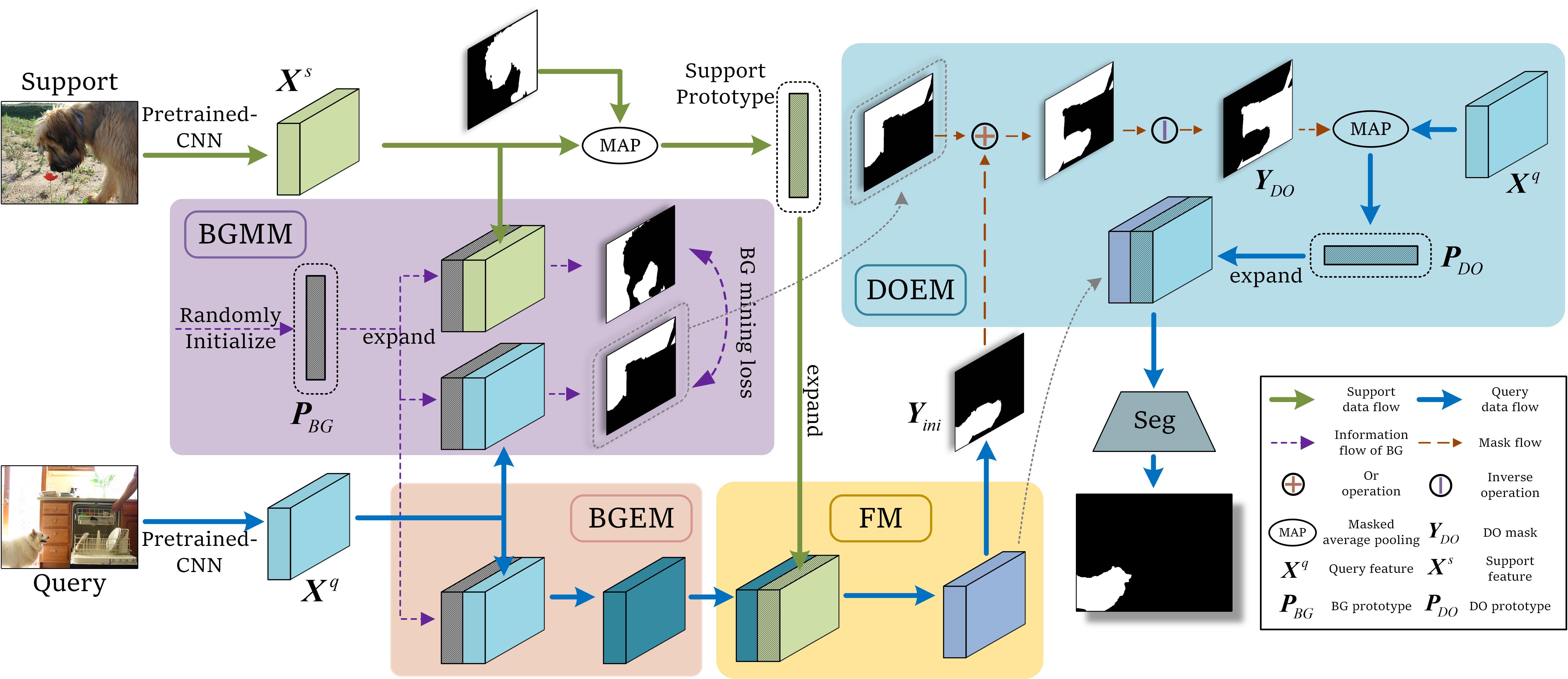}
	\end{center}
	\setlength{\abovecaptionskip}{0.cm}
	\caption{Overall architecture of the proposed method for few-shot semantic segmentation. Our network is composed of four parts. After extracting features from both the support and query images via a pre-trained backbone, our Background Mining Modul (BGMM) is performed to obtain a BG prototype and segment the BG regions. Meanwhile, Background Eliminating Module (BGEM) is performed to eliminate the BG regions. The third part is to obtain the activated query feature and further an initial target prediction via Feature Matching (FM). The last part is to eliminate the distracting objects by our proposed Distracting Objects Eliminating Module (DOEM).}
	\label{fig:short}
	\vspace{-3mm}
\end{figure*}

\vspace{-1mm}
\section{Proposed Method}
\vspace{-1mm}
\subsection{Problem Definition}
Few-shot semantic segmentation aims to train a model on base classes, and segment unseen objects in query images with a few annotated support samples without re-training. Typically, all the datasets are divided into two subsets. One is the training set \large$ \mathcal D_{base}$ \normalsize with the base classes \large$ \mathcal C_{base}$\normalsize. The other is the testing set \large$ \mathcal D_{novel}$ \normalsize with the novel classes \large$ \mathcal C_{novel}$\normalsize. 
These two sets of classes are disjoint, \ie, \large $ \mathcal C_{base}\cap \mathcal C_{novel}=\varnothing  $\normalsize. 
Specifically, the training set \large$ \mathcal D_{base}$ \normalsize is partitioned into several episodes after randomly sampling $K+1$ image-mask pairs that contain objects from a specific class in \large$ \mathcal C_{base}$\normalsize. The testing set is composed of similar episodes, except that the data are sampled from the \large$ \mathcal C_{novel}$\normalsize. For one episode, $K$ image-mask pairs are treated as the support set \large $ \mathcal S$\normalsize$=\{(\bm{I}^{s}_{i},\bm{M}^{s}_{i})\}^{K}_{i=1} $ to segment the objects of the target class in the remaining one sample, which is termed the query set \large$ \mathcal Q $\normalsize.
Here, $\bm{I} \in \mathbb{R}$$^{H\times W\times 3}$ indicates the RGB image and $\bm{M} \in \mathbb{R}$$^{H\times W}$ indicates the corresponding mask. \large $ \mathcal Q$\normalsize$=\{(\bm{I}^{q},\bm{M}^{q})\} $ is provided with the ground truth only during training. Following this episode, the network is trained on \large$ \mathcal D_{base}$ \normalsize and evaluated on \large$ \mathcal D_{novel}$ \normalsize.
\vspace{-1mm}
\subsection{Overview}
As aforementioned, few-shot semantic segmentation models usually fail in ambiguous non-target regions. Motivated by this
observation, we propose to mine and eliminate non-target regions, which include the background (BG) regions and the distracting object (DO) regions.

We first follow previous methods and use a pre-trained backbone to extract the query feature map $\bm{X}^{s}\in \mathbb{R}$$^{H\times W\times C}$ and the support feature map $\bm{X}^{q}\in \mathbb{R}$$^{H\times W\times C}$ from corresponding images, respectively.
Then, the BG mining module (BGMM) is proposed to mine the BG region via learning a general BG prototype, which is randomly initialized and subsequently learned on the training set. We also propose a novel BG loss without using accurate BG segmentation ground truth.
Next, a BG eliminating module (BGEM) is utilized to filter out the BG information from the query feature. Subsequently, we follow prototypical learning to activate the target region in the query feature using the support target prototype and feature matching (FM). An initial target object segmentation mask can be further obtained.

After that, a DO eliminating module (DOEM) is proposed to filter out DO information from the query feature. The DO region can be first mined by combining the BG segmentation map and the initial target prediction. Then, the DO prototype can be obtained from the query feature and used for DO elimination. The prototypical contrastive learning (PCL) is also adopted for better discriminating the target object from DOs.
Finally, a segmentation network is used to achieve the BG and DO-free prediction. 
\vspace{-1mm}
\subsection{Background Mining and Eliminating}
\subsubsection{Background Mining Module}
Background regions, in which no obvious objects appear, commonly exist in most images. Based on this commonality, we propose to use a BG prototype to encode general BG knowledge. It can be represented as $\bm{P}_{BG} \in \mathbb{R}$$^{1\times 1\times D}$, where $D$ is the channel dimension.
Inspired by some saliency detection methods \cite{liu2018picanet,liu2021visual}, it is feasible to learn $\bm{P}_{BG}$ on a large number of images and then use it to detect BG regions on any natural image.

Hence, in this paper, we first randomly initialize $\bm{P}_{BG}$ and then learn it from both support and query images in the base classes during training. Specifically, given the query feature map $\bm{X}^{Q}$ and the support feature map $\bm{X}^{S}$ extracted from the backbone, 
we expand $\bm{P}_{BG}$ to the same size as them, obtaining $\bm{\hat{P}}_{BG} \in \mathbb{R}$$^{H\times W\times C}$. Then, we concatenate it with $\bm{X}^{q}$ and $\bm{X}^{s}$, respectively, and use a simple segmentation network $\mathcal{F}_{3\times3}(\cdot)$
to get the BG prediction for the query and the support:
\begin{eqnarray}
\setlength{\abovedisplayskip}{3pt}
\setlength{\belowdisplayskip}{3pt}
\label{eq:bgmm}
	\bm{y}_{BG}^q=\mathcal{F}_{3\times3}(\bm{X}^{q}\oplus \bm{\hat{P}}_{BG}),\\
\label{eq:bgmm2}
	\bm{y}_{BG}^s=\mathcal{F}_{3\times3}(\bm{X}^{s}\oplus \bm{\hat{P}}_{BG}),
\end{eqnarray}
where $\oplus$ denotes the concatenation operation along the channel dimension and $\mathcal{F}_{3\times3}$ is composed of two $3\times3$ convolutional layers and shares the same weights in both \eqref{eq:bgmm} and \eqref{eq:bgmm2}. $\bm{y}_{BG}^{q/s}\in \mathbb{R}$$^{H\times W\times 1}$ is the BG segmentation probability map of the query or the support.

\vspace{-4mm}
\paragraph{Background Mining Loss.}
In the few-shot semantic segmentation task, we only have the ground truth of target object masks, \ie, $\bm{M}^{q}$ and $\bm{M}^{s}$, and have no BG region ground truth. In order to force $\bm{P}_{BG}$ effectively predict the BG region as we expected, we design a BG mining loss to optimize this learning process as below:
\begin{equation}
\label{eq:bg_loss}
\setlength{\abovedisplayskip}{3pt}
\setlength{\belowdisplayskip}{3pt}
\begin{split}
    L_{BG}=&-\frac{1}{N}\sum_{i}log(1-\bm{y}_{BG}^{q/s}(i))\bm{M}^{q/s}(i)\\
    &-\alpha \frac{1}{Z}\sum_{j}log(\bm{y}_{BG}^{q/s}(j)),
\end{split}
\end{equation}
where $i$ and $j$ are the indexes of the spatial locations. $\bm{M}^{q/s}$ is the ground truth of target objects belonging to query or support. $N$ denotes the total number of the target object pixels and $Z$ is equal to $H\times W$.
$\alpha$ is a hyperparameter to weight the second term.

The core idea of this loss is that the BG prediction should belong to the reverse region of the target object, \ie, predicting zero in $\bm{y}_{BG}^{q/s}$ for the pixels that belong to the target object. However, solely using this constraint may lead to a trivial solution that predicts all zeros for $\bm{y}_{BG}^{q/s}$.
To alleviate this problem, we add the second term as a regularization to force the module must predict valid BG regions for every image.
\vspace{-3mm}
\subsubsection{Background Eliminating Module}
We further use the expanded BG prototype $\bm{\hat{P}}_{BG}$ to filter out the BG information from the query feature map via prototypical learning. Specifically, we first concatenate $\bm{\hat{P}}_{BG}$ with $\bm{X}^{q}$ and then use a convolutional layer $\mathcal{F}_{1\times1}(\cdot)$ to exclude the BG information in the query. The whole process can be denoted as:
\begin{equation}
\setlength{\abovedisplayskip}{3pt}
\setlength{\belowdisplayskip}{3pt}
\label{eq:bgem}
	\bm{X}_{BG}^{q}=\mathcal{F}_{1\times1}(\bm{X}^{q}\oplus \bm{\hat{P}}_{BG}),
\end{equation}
where $\bm{X}_{BG}^{q}\in \mathbb{R}$$^{H\times W\times C}$ denotes the BG-filtered query feature and $\mathcal{F}_{1\times1}(\cdot)$ denotes a 1$\times$1 convolutional layer.

\subsection{Support Feature Matching}\label{sec:sfm}
Following previous methods, we further use dense feature matching to activate the target object region on the $\bm{X}_{BG}^{q}$.
Concretely, masked average pooling (MAP) is first used on the support feature map $\bm{X}^{s}$ to get the support prototype $\bm{P}^{s}\in \mathbb{R}$$^{1\times 1\times C}$. Then, it is expanded to $\bm{\hat{P}}^{s}\in \mathbb{R}$$^{H\times W\times C}$ and concatenated with the BG-filtered query feature $\bm{X}_{BG}^{q}$. We also follow \cite{tian2020prior} and introduce a prior confidence map $\bm{C}_{p}\in \mathbb{R}$ $^{H\times W\times 1}$ via computing the maximum similarity score at pixel-level. After that, we obtain the activated query feature $\bm{X}_{act}^{q}\in \mathbb{R}$$^{H\times W\times C}$ and further achieve the initial target object prediction $\bm{y}_{ini}^{q}\in \mathbb{R}$$^{H\times W\times 1}$:
\begin{eqnarray}
\setlength{\abovedisplayskip}{3pt}
\setlength{\belowdisplayskip}{3pt}
	&\bm{X}_{act}^{q}=\mathcal{F}_{1\times1}(\bm{X}_{BG}^{q}\oplus \bm{\hat{P}}^{s}\oplus \bm{C}_{p}),\\
	&\bm{y}_{ini}^{q}=\mathcal{F}_{3\times3}(\bm{X}_{act}^{q}),
\end{eqnarray}
where $\mathcal{F}_{1\times1}(\cdot)$ is the same as in \eqref{eq:bgem} and $\mathcal{F}_{3\times3}(\cdot)$ is the same as in \eqref{eq:bgmm}.

\subsection{Distracting Objects Eliminating}

\subsubsection{Distracting Object Eliminating Module}
Although we have eliminated BG information in $\bm{X}_{act}^{q}$, it may still suffer from the distraction of DOs. To this end, we design the DOEM to further filter out DO information from $\bm{X}_{act}^{q}$ for more accurate target object prediction. 
To be specific, we mine the potential DO region in the query based on the known BG region in $\bm{y}_{BG}^{q}$ and the target object region in $\bm{y}_{ini}^{q}$. Intuitively, the DO region is complementary to the union of the BG region and target region. Hence, we have:
\begin{equation}
	\bm{Y}_{DO}^{q}=1-(\bm{Y}_{BG}^{q} \cup \bm{Y}_{ini}^{q}),
\end{equation}
where $\bm{Y}_{DO} \in \mathbb{R}$$^{H\times W\times 1}$ denotes the DO mask. $\bm{Y}_{BG}^{q}$ and $\bm{Y}_{ini}^{q}$ are the binary maps corresponding to $\bm{y}_{BG}^{q}$ and $\bm{y}_{ini}^{q}$, respectively.

Next, we utilize $\bm{Y}^{q}_{DO}$ to obtain the DO prototype $\bm{P}_{DO}^{q} \in \mathbb{R}$$^{1\times 1\times C}$ via performing MAP on the query feature map:
\begin{equation}
\setlength{\abovedisplayskip}{3pt}
\setlength{\belowdisplayskip}{3pt}
\label{eq:do_proto}
	\bm{P}_{DO}^{q}=\frac{\sum\bm{X}^{q} \otimes   \bm{Y}_{DO}^{q}}{\sum\bm{Y}_{DO}^{q}},
\end{equation}
where $\otimes$ denotes the element-wise multiplication and the summation sums over all spatial locations.

After that, we expand the $\bm{P}_{DO}^{q}$ into $\bm{\hat{P}}_{DO}^{q}\in \mathbb{R}$$^{H\times W\times C}$ and combine it with the activated query feature map $\bm{X}_{ini}^{q}$ to eliminate the DO information. Finally, the combined feature is passed into a segmentation network, for which we use the Feature Enrichment Module (FEM) in \cite{tian2020prior}, to obtain the BG and DO-free prediction: 
\begin{equation}
\setlength{\abovedisplayskip}{3pt}
\setlength{\belowdisplayskip}{3pt}
	\bm{y}^{q} = \text{Seg}(\bm{X}_{ini}^{q}\oplus \bm{\hat{P}}_{DO}^{q}),
\end{equation}
where $\bm{y}^{q}\in \mathbb{R}$$^{H\times W\times 1}$ is the final target object segmentation result of our whole model.

\vspace{-3mm}
\subsubsection{Prototypical Contrastive Learning}
The DOEM only cares about the DO mask $\bm{Y}_{DO}^{q}$ in the DO eliminating process. However, a good DO eliminating model requires not only accurate DO masks, but also good prototype feature embeddings that can differentiate the target objects from DOs easily.
Inspired by recent research on contrastive learning, we propose the prototypical contrastive learning (PCL) method to refine the feature embeddings of different prototypes. With the help of PCL, we want to make the prototype features between the target objects and the DOs more discriminative, and the prototypes between the target objects of the query and the support more similar.

To this end, we need to obtain the target prototypes and DO prototypes for both the query and the support first. For the target prototypes, we have obtained it for the support, \ie, $ \bm{P}^{s}$, from Section~\ref{sec:sfm}. For the query image, we first binarize the final target prediction $\bm{y}^{q}$ as the target mask and then adopt MAP on the query feature to generate the target prototype of the query $\bm{P}^{q}\in \mathbb{R}$$^{1\times 1\times C}$. As for the DO prototypes, we have computed it for the query, \ie, $\bm{P}_{DO}^{q}$, in \eqref{eq:do_proto}. For the support image, we adopt the same workflow to compute the DO prototype of the support $\bm{P}_{DO}^{s}$, \ie, using the target mask $\bm{M}^{s}$ and the BG mask $\bm{Y}_{BG}^{s}$ to generate the DO mask, and then conducting MAP on the support feature. 


\vspace{-3mm}
\paragraph{Prototypical Contrastive Learning Loss.}
According to the paradigm of contrastive learning, we propose a PCL loss to optimize the above prototype feature embeddings. For the query prototype $\bm{P}^{q}$, we treat the corresponding support prototype $ \bm{P}^{s}$ as the positive sample, while the DO prototypes in both query and support as negative samples. Considering that a large number of negative samples is indispensable for contrastive learning, we build a DO prototype bank $\mathcal{B}$ to store the embeddings of 2000 DO prototypes in the latest batches during training.
Note that they can be sampled across different episodes of the same class.
At last, inspired by InfoNCE\cite{oord2018representation}, we propose our PCL loss:
\begin{align}
\label{eq:pcl_loss}
\begin{split}
	L_{PCL}\! =\!
	-log \frac{e^{cos(\bm{P}^{q},\bm{P}^{s})}}
	{\sum_{\mathcal{B}} \{e^{cos(\bm{P}^{q}, \bm{P}_{DO}^{q})} + e^{cos(\bm{P}^{q}, \bm{P}_{DO}^{s})}\}},
\end{split}
\end{align}
where $cos(,)$ denotes the cosine similarity.


\subsection{Total Training Loss}
We use two binary cross-entropy losses to supervise the training of the initial target prediction $\bm{y}_{ini}^{q}$ and the final prediction $\bm{y}^{q}$, composing the target segmentation loss $L_{T}$.
Finally, our total training loss includes $L_{T}$, the BG loss $L_{BG}$ in \eqref{eq:bg_loss}, and the PCL loss $L_{PCL}$ in \eqref{eq:pcl_loss}:
\begin{equation}
\setlength{\abovedisplayskip}{3pt}
\setlength{\belowdisplayskip}{3pt}
	L= \beta L_{T}+\lambda L_{BG}+\gamma L_{PCL},
\end{equation}
\vspace{-3mm}
\begin{equation}
\setlength{\abovedisplayskip}{3pt}
\setlength{\belowdisplayskip}{3pt}
	L_{T}= \text{BCE}(\bm{y}_{ini}^{q},\bm{M}^q )+\text{BCE}(\bm{y}^{q},\bm{M}^q ),
\end{equation}
where BCE denotes the binary cross-entropy loss and $\beta, \lambda, \gamma$ are adjustable loss weights.

\section{Experiments}

\subsection{Datasets and Evaluation Metrics}

\paragraph{Datasets.}
We evaluate our model on two benchmark datasets, \ie, the PASCAL-$5^{i}$ dataset\cite{shaban2017one} and the COCO-$20^{i}$ dataset \cite{nguyen2019feature}. PASCAL-$5^{i}$ is constructed based on the PASCAL VOC 2012 dataset \cite{everingham2010pascal} and external annotations from SDS \cite{hariharan2011semantic}. The total 20 categories are partitioned into 4 folds as in \cite{wang2019panet} for cross validation and each fold contains 5 categories. COCO-$20^{i}$ is a larger datasets based on the MSCOCO\cite{lin2014microsoft} dataset. Similar to PASCAL-$5^{i}$, the total 80 categories are also partitioned into 4 folds for cross validation, where each fold includes 20 categories. For both the datasets, we test on 1 fold and train on the remaining 3 folds.

\vspace{-4mm}
\paragraph{Evaluation Metrics.}
Following previous methods \cite{shaban2017one,siam2019amp,liu2020crnet,liu2020part}, we adopt the class mean intersection over union (mIoU) as a primary evaluation metric for ablation studies and comparisons. In addition, we report the results of foreground-background IoU (FB-IoU), which only cares about the performance on target and non-target regions instead of differentiating categories, for a more comprehensive comparison. , the precision, whose formulation follows $\frac{TP}{TP+FP}$, also be leveraged to report our modules performance on decreasing false positives.

\begin{figure}[t]
	\centering
	\includegraphics[scale=0.28]{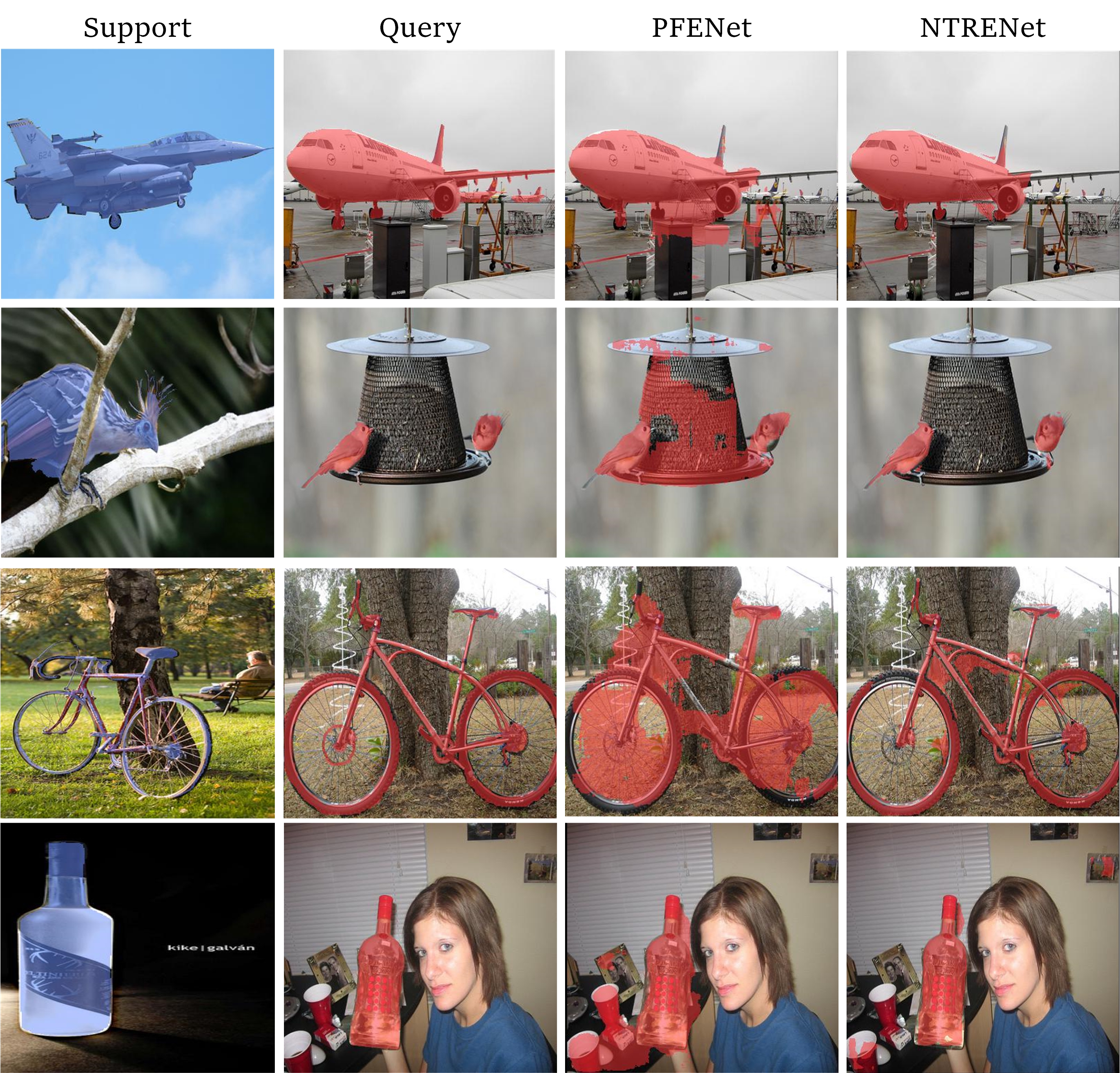}
	\vspace{-3mm}
	\caption{Qualitative results of our proposed NTRENet and PFENet. From left to right: support images, query images, prediction of PFENet, prediction of NTRENet.}
	\label{figurecompar}
	\vspace{-3mm}
\end{figure}

\subsection{Implementation Details}
Following previous works, we respectively use ResNet-50, ResNet-101 \cite{he2016deep}, and VGG-16 \cite{simonyan2014very} as the backbones to construct our network for fair comparisons. These backbones are all pre-trained on the ImageNet classification task and their weights are fixed during training.

\begin{table*}[ht]
	\centering
	\caption{Class mIoU and FB-IoU results of four folds on PASCAL-$5^{i}$. The results of ‘Mean’ are the averaged class mIoU scores of all four folds. The detailed FB-IoU results of each fold are omitted in this table for simplicity. \textbf{Bold} indicates the best results.}
	\label{Voccompar}
	\vspace{-3mm}
	\resizebox{160mm}{40mm}{
		\begin{tabular}{c|l|cccc|cc|cccc|cc}
			\hline
			\multirow{2}{*}{Backbone}   & \multicolumn{1}{c|}{\multirow{2}{*}{Methods}} & \multicolumn{6}{c|}{1-Shot}       & \multicolumn{6}{c}{5-Shot}                                \\ \cline{3-14} 
			& \multicolumn{1}{c|}{}                         & Fold-0 & Fold-1 & Fold-2 & Fold-3 & \textbf{Mean} & FB-IoU & Fold-0 & Fold-1 & Fold-2 & Fold-3 & \textbf{Mean} & FB-IoU \\ \hline
			\multirow{6}{*}{VGG-16}    & OSLSM  \cite{shaban2017one}                                       & 33.6   & 55.3   & 40.9   & 33.5   & 40.8          & 61.3   & 35.9   & 58.1   & 42.7   & 39.1   & 44.0          & 61.5   \\
			& co-FCN     \cite{rakelly2018conditional}                                   & 36.7   & 50.6   & 44.9   & 32.4   & 41.1          & 60.1   & 37.5   &50.0     & 44.1   & 33.9   & 41.4          & 60.2   \\
			& RPMM  \cite{yang2020prototype}                                       & 47.1   & 65.8   & 50.6   & 48.5   & 53.0          & -      & 50.0   & 66.5   & 51.9   & 47.6   & 54.0          & -      \\
			& PFENet\cite{tian2020prior}                                          & 56.9   & \textbf{68.2}   & 54.4   & 52.4   & 58.0          & 72.3   & 59.0   & \textbf{69.1}   & 54.8   & 52.9   & 59.0          & 72.3   \\ 
			& MMNet\cite{wu2021learning} 
			& 57.1  & 67.2 & 56.6  & 52.3   & 58.3  & - 
			        & 56.6   & 66.7   & \textbf{63.6}   & 56.5  & 58.3   & -  
			        \\\cline{2-14}
			& NTRENet                                        & \textbf{57.7}   &67.6&\textbf{57.1 }   &\textbf{53.7}    &\textbf{59.0}           &\textbf{73.1}    & \textbf{60.3}   &68.0    &55.2    &\textbf{57.1}    &\textbf{60.2}         & \textbf{74.2}  \\ \hline
			\multirow{9}{*}{ResNet-50} & CANet   \cite{zhang2019canet}                                      & 52.5   & 65.9   & 51.3   & 51.9   & 55.4          & 66.2   & 55.5   & 67.8   & 51.9   & 53.2   & 57.1          & 69.6   \\
			& RPMM \cite{yang2020prototype}                                         & 55.2   & 66.9   & 52.6   & 50.7   & 56.3          & -      & 56.3   & 67.3   & 54.5   & 51.0   & 57.3          & -      \\
			& PFENet\cite{tian2020prior}                                          & 61.7  & 69.5   & 55.4   & 56.3     & 60.8          & 73.3   & 63.1   & 70.7   & 55.8   & 57.9   & 61.9          & 73.9   \\
			& SCL\cite{zhang2021self}                                            & 63.0   & 70.0   & 56.5   & 57.7   & 61.8          & 71.9   & 64.5   & 70.9   & 57.3   & 58.7   & 62.9          & 72.8   \\
			& ASGNet\cite{li2021adaptive}                                       & 58.8   & 67.9   & 56.8   & 53.7   & 59.3          & 69.2   & 63.7   & 70.6   & 64.2   & 57.4   & 63.9          & 74.2   \\
			& ReRPI\cite{boudiaf2021few}                   
			& 59.8   & 68.3   & 62.1   & 48.5   & 59.7          & -      & 64.6   & 71.4   & 71.1   & 59.3   &\textbf{66.6}          & -      \\ 
			& SAGNN\cite{xie2021scale}                   
			& 64.7  & 69.6   & 57.0   & 57.2   & 62.1          & 73.2   & 64.9   & 70.0   & 57.0 &59.3   & 62.8          & 73.3  \\ 
			& MLC\cite{yang2021mining}                                        
			& 59.2 & 71.2   & \textbf{65.6}   & 52.5   & 62.1          & -      
			& 63.5   & 71.6   & \textbf{71.2}   & 58.1   & 66.1          & -      \\  \cline{2-14}
			& NTRENet                                        & \textbf{65.4} & \textbf{72.3} & 59.4   &  \textbf{59.8}    & \textbf{64.2}          &  \textbf{77.0}      & \textbf{66.2}   & \textbf{72.8}   & 61.7 & \textbf{62.2}   & 65.7          &\textbf{ 78.4 }      \\ \hline

			\multirow{8}{*}{ResNet-101} 
			& DAN \cite{wang2020few}                                           & 54.7   & 68.6   & 57.8   & 51.6   & 58.2          & 71.9   & 57.9   & 69.0   & 60.1   & 54.9   & 60.5          & 72.3   \\
			& PPNet \cite{liu2020part}                                          & 52.7   & 62.8   & 57.4   & 47.7   & 55.2          & 70.9   & 60.3   & 70.0   & 69.4   & 60.7   & 65.1          & 77.5   \\
			& PFENet\cite{tian2020prior}                                          & 60.5   & 69.4   & 54.4   & 55.9   & 60.1          & 72.9   & 62.8   & 70.4  & 54.9   & 57.6   & 61.4          & 73.5   \\
			& ASGNet\cite{li2021adaptive}                                         & 59.8   & 67.4   & 55.6   & 54.4   & 59.3          & 71.7   & 64.6   & 71.3   & 64.2   & 57.3   & 64.4          & 75.2   \\
			& ReRPI\cite{boudiaf2021few}                                         & 59.6   & 68.6   & \textbf{62.2}   & 47.2   & 59.4          & -      & 66.2   & 71.4   & 67.0   & 57.7   & 65.6          & -      \\  
			& MLC\cite{yang2021mining}                                        
			& 60.8 & 71.3   & 61.5   & 56.9   & 62.6          & -      
			& 65.8   & \textbf{74.9}   & \textbf{71.4}   & 63.1   &\textbf{68.8}          & -      \\  \cline{2-14}
			& NTRENet                                        &\textbf{65.5}    &\textbf{71.8}    &59.1  &\textbf{58.3}    &\textbf{63.7}           & \textbf{75.3}   & \textbf{67.9}       &    73.2&60.1    &\textbf{66.8}    & 67.0    & \textbf{78.2}       \\ \hline
	\end{tabular}}
\end{table*}
\begin{table*}[]
	\centering
	\caption{Class mIoU and FB-IoU results of four folds on COCO-$20^{i}$. The results of ‘Mean’ are the averaged class mIoU scores of all the four folds. The detailed FB-IoU results of each fold are omitted in this table for simplicity. \textbf{Bold} indicates the best results.
	}
	\label{Cococompar}
	\vspace{-3mm}
	\resizebox{160mm}{!}{
		\begin{tabular}{l|l|cccc|cc|cccccc}
			\hline
			\multicolumn{1}{c|}{\multirow{2}{*}{Backbone}}   & \multicolumn{1}{c|}{\multirow{2}{*}{Methods}} & \multicolumn{6}{c|}{1-Shot}                                                                                                                                   & \multicolumn{6}{c}{5-Shot}                                                                                                                                     \\ \cline{3-14} 
			\multicolumn{1}{c|}{}                            & \multicolumn{1}{c|}{}                         & Fold-0                   & Fold-1                   & Fold-2                   & Fold-3                    & \textbf{Mean}            & FB-IoU                & Fold-0                   & Fold-1                   & Fold-2                   & \multicolumn{1}{c|}{Fold-3} & \textbf{Mean}            & FB-IoU               \\ \hline

			\multirow{6}{*}{ResNet-50}                       & PPNet \cite{liu2020part}                                        & 28.1                     & 30.8                     & 29.5                     & 27.7                      & 29.0                     & -                     & 39.0                     & 40.8                     & 37.1                     & \multicolumn{1}{c|}{37.3}   & 38.5                     & -                    \\
			& RPMM  \cite{yang2020prototype}                                        & 29.5                     & 36.8                     & 28.9                     & 27.0                      & 30.6                     & -                     & 33.8                     & 42.0                     & 33.0                     & \multicolumn{1}{c|}{33.3}   & 35.5                     & -                    \\

			& ASGNet\cite{li2021adaptive}                                         & -  & -  & - & -  & 34.6  & 60.4  & - & -   & - 
			& \multicolumn{1}{c|}{-} & \textbf{42.5}    & 67.0   \\
			& MMNet\cite{wu2021learning}
			& 34.9  & 41.0 & 37.2  & 37.0   & 37.5  & - 
			        & 37.0   & 40.3   & 39.3   & \multicolumn{1}{c|}{36.0}  & 38.2   & -  
			        \\
			& MLC\cite{yang2021mining}                                        
			& \textbf{46.8} & 35.3   & 26.2   & 27.1   & 33.9          & -      
			& \textbf{54.1}   & 41.2   & 34.1   & \multicolumn{1}{c|}{33.1}   & 40.6          & -      \\  \cline{2-14}
			& NTRENet                                        &    36.8                      & \textbf{42.6}    & \textbf{39.9}                         &  \textbf{37.9}    &  \textbf{39.3}                  &\textbf{68.5}               &38.2                    &\textbf{44.1}   &\textbf{40.4}                   & \multicolumn{1}{c|}{\textbf{38.4}}       & 40.3                     & \textbf{69.2}                \\ \hline
			\multicolumn{1}{c|}{\multirow{6}{*}{ResNet-101}} 
                       & DAN \cite{wang2020few}                                          & -                        & -                        & -                        & -                         & 24.4                     & 62.3                  & -                        & -                        & -                        & \multicolumn{1}{c|}{-}      & 29.6                     & 63.9                 \\
			\multicolumn{1}{c|}{}                            & SCL\cite{zhang2021self}                                            & 36.4                     & 38.6                     & 37.5                     & 35.4                      & 37.0                     & -                     & 38.9                     & 40.5                     & 41.5                     & \multicolumn{1}{c|}{38.7}   & 39.9                     & -                    \\
			\multicolumn{1}{c|}{}                            & PFENet\cite{tian2020prior}                                        & 34.3                     & 33.0                     & 32.3                     & 30.1                      & 32.4                     & 58.6                  & 38.5                     & 38.6                     & 38.2                     & \multicolumn{1}{c|}{34.3}   & 37.4                     & 61.9                 \\
			& MLC\cite{yang2021mining}                            
			& \textbf{50.2} & 37.8   & 27.1   & 30.4   & 36.4          & -      
			& \textbf{57.0}   & 46.2   & 37.3   & \multicolumn{1}{c|}{37.2}   & \textbf{44.4}         & -      \\ 
			& SAGNN\cite{xie2021scale}         
			& 36.1 &\textbf{41.0}   & 38.2   & 33.5   & 37.2          & 60.9  
			& 40.9   & \textbf{48.3}   & 42.6   & \multicolumn{1}{c|}{38.9}   & 42.7          & 63.4   \\ \cline{2-14}
			\multicolumn{1}{c|}{}                            & NTRENet                                        &  38.3 &40.4                    & \textbf{ 39.5 }                   &\textbf{38.1 }                      &\textbf{39.1}                     & \textbf{67.5}                 &42.3                      &  44.4                    & \textbf{44.2}                   & \multicolumn{1}{c|}{\textbf{41.7}}   &43.2                      & \textbf{69.6} \\ \hline
	\end{tabular}}
	\vspace{-3mm}
\end{table*}
\begin{figure*}[t]
	\begin{center}
		\includegraphics[scale = 0.29]{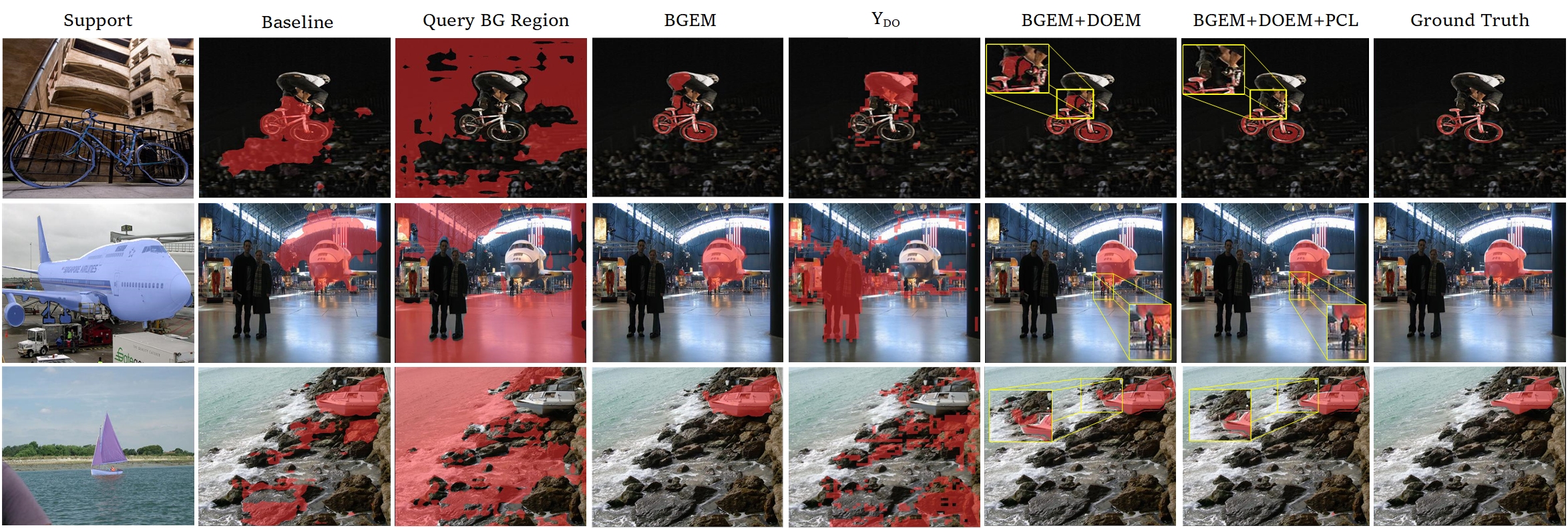}
	\end{center}
	\setlength{\abovecaptionskip}{0.cm}
	\vspace{-3mm}
	\caption{Visualization of different ablative results. From left to right: Support images, the results of baseline, BG prediction, the results of only using BGEM, DO prediction, the results of using BGEM+DOEM, the results of using BGEM+DOEM+PCL (\ie, full model), Ground truth. }
	\label{ablation_pic}
\end{figure*}
Our network is implemented using PyTorch\cite{paszke2019pytorch} and all the experiments are conducted on one NVIDIA RTX 3090 GPU.
We use random scaling, horizontal flipping, and random rotation within [-10,10] degrees as data augmentation to increase the training data. Finally, we randomly crop images and masks with the size of $473\times473$ to train our model.
During training, we use SGD as our optimizer, where the initial learning rate, batch size, weight decay, and momentum are set as 0.03, 32, 0.0001, and 0.9, respectively. We train our model for 200 epochs on PASCAL-$5^{i}$ and 50 epochs on COCO-$20^{i}$, respectively. The learning rate is decayed using the polynomial annealing policy with the power set to 0.9.
During the evaluation, we follow \cite{yang2021mining} to randomly sample 1000 support-query pairs on PASCAL-$5^{i}$ and 4000 pairs on COCO-$20^{i}$, respectively.

\subsection{Comparison with State-of-the-art Methods}
\paragraph{PASCAL-$5^{i}$.} 
Table \ref{Voccompar} shows the performance comparison on PASCAL-$5^{i}$ between our method and several representative models. We can see that, on all the three backbones(\ie, VGG-16, Resnet-50, and Resnet-101), our method outperforms all previous models by a large margin in terms of both mIoU and FB-IoU. Specifically, under the 1-shot setting, the averaged mIoU scores of our method are 59.0, 64.2, and 63.7 on the VGG-16, Resnet-50, and Resnet-101 backbones, respectively, surpassing state-of-the-art results by 1.2\%, 3.4\%, and 1.8\%, respectively. Meanwhile, in terms of FB-IoU, our method outperforms the previous best results by 1.1\%, 5.0\%, and 3.3\% on the three backbones, respectively. Under the 5-shot setting, our method only obtains the best results on the VGG-16 backbone in terms of mIoU, but outperforms previous state-of-the-art FB-IoU results by 2.6\%, 5.7\%, and 1\% on all three backbones, respectively.
\vspace{-4mm}
\paragraph{COCO-$20^{i}$.}
Although COCO-$20^{i}$ is a more challenging dataset with a large number of images with realistic scenes, we still obtain superior performance, which is shown in Table \ref{Cococompar}. Here we follow previous works and only use the Resnet-50 and Resnet-101 backbones. Table \ref{Cococompar} shows that our method respectively yields the averaged mIoU scores of 39.3 and 39.1 on the two backbones under the 1-shot setting, outperforming previous best results by a large margin of 4.8\% and 5.7\%, respectively. In the 5-shot setting, the FB-IoU results also verify the superiority of our method, despite the challenging scenarios.
\vspace{-4mm}
\paragraph{Limitations.}
We find that the averaged mIoU results of our model do not achieve obvious advantages in the 5-shot setting, compared with previous state-of-the-arts.
We argue that this is reasonable since our method mainly focuses on eliminating non-target regions instead of segmenting the target object. As such, although the number of support samples increases from one to five, it does not introduce additional non-target information to our method. 
However, we hope our work could provide a novel perspective on the opposite side of traditional methods for future works.
\vspace{-4mm}
\paragraph{Qualitative Comparison.} 
We show the qualitative comparison of the predicted segmentation masks generated by our method and a typical traditional model that focuses on segmenting the target object, \ie, PFENet \cite{tian2020prior}, in Figure~\ref{figurecompar}. We can see that PFENet could not segment the target objects accurately due to the distraction of non-target regions.
However, our proposed NTRENet can obtain much more accurate results with much fewer false positive predictions in BG and DO regions, thus clearly demonstrating the effectiveness of our proposed method.
\vspace{-1mm}
\subsection{Ablation Study}
\vspace{-1mm}
\paragraph{Effectiveness of Different Modules.} We conduct extensive ablation studies on PASCAL-${5^{i}}$ in the 1-shot setting to validate the effectiveness of our proposed key modules, \ie, BGEM, DOEM, and PCL. We remove these three modules from our NTRENet as the baseline model, which only uses the support prototype to directly segment the target object as in \cite{tian2020prior}.
As Table~\ref{abulation} shows, eliminating the BG regions using BGEM achieves as large as 4\% performance improvement compared to the baseline model. Meanwhile, using DOEM to mine and eliminate DO regions obtains another 2\% performance gain. Finally, using the PCL scheme to boost the model capability of discriminating different objects leads to further 1\% performance improvement. These results clearly demonstrate the effectiveness of our proposed BGEM, DOEM, and PCL. In addition, we use precision to verify the performance of our method on decreasing false positives. The results show that our full method achieves 3\% precision improvement compared to the baseline model and all of the proposed BGEM, DOEM, and PCL can progressively improve the precision, \ie, decrease false positives.  
\vspace{-3mm}
\begin{table}[t]
	\centering
	\caption{Ablation study of the key modules in our NERTNet. mIoU results are reported on the PASCAL-${5^{i}}$ dataset under the 1-shot setting.
	}
	\label{abulation}
	\resizebox{83mm}{8mm}{
	\begin{tabular}{ccc|cccccc}
			\hline
			\multicolumn{1}{c}{BGEM} & \multicolumn{1}{c}{DOEM} & \multicolumn{1}{c|}{PCL} & Fold-0 & Fold-1 & Fold-2 & Fold-3 & \textbf{Mean} &Precision\\ \hline
			&     &                                                                        & 60.8  &68.2  &   55.4   & 55.3      & 60.0    & 61.9 \\
			\multicolumn{1}{c}{$\checkmark$}            &                         &                                                                        & 63.2    & 71.1	&57.7& 57.4       &62.4     &62.8 \\
			\multicolumn{1}{c}{$\checkmark$}          &    \multicolumn{1}{c}{$\checkmark$}           &                                                                        &   64.7 &    71.9&58.8    &59.0    &63.6 &63.3 \\

			\multicolumn{1}{c}{$\checkmark$}            &   \multicolumn{1}{c}{$\checkmark$}          &    \multicolumn{1}{c|}{$\checkmark$}                                                    & \textbf{65.4}       & \textbf{72.3}   &\textbf{59.4}   &\textbf{59.8}    & \textbf{64.2} &\textbf{63.6}\\ \hline
	\end{tabular}}
		\vspace{-3mm}
\end{table}

\vspace{-4mm}
\paragraph{Choice of negative samples in PCL.}
The whole non-target
region includes both BG and DO regions. Hence, we choose other alternative negative samples, \ie, prototypes generated from the whole non-target region and the BG region, respectively, to apply in PCL. We conduct comparative experiments on PASCAL-${5^{i}}$ under the 1-shot setting. In Table~\ref{abl_pcl}, the results show that DO prototypes work more effectively than others since DO regions are more confusing and thus play a role of hard negative samples. 

\vspace{-5mm}
\paragraph{Qualitative Comparison.}
We further show some qualitative results in Figure~\ref{ablation_pic} to prove the effectiveness of our proposed BGEM, DOEM, and PCL in an intuitionistic way. 
Column 2 shows predictions from baseline. In column 3, we show the BG prediction masks obtained from BGMM. We find that our BGMM can effectively mine the universal BG regions in query. Column 5 reveals the predicted masks of DO regions in DOEM. Columns 4 and 6 show that using BGEM and DOEM can effectively help eliminate the BG and DO regions compared with the baseline results. Finally, column 7 indicates that using PCL can further discriminate the target objects from DOs in detail.

\begin{table}[t]
	\centering
	\caption{Comparison of the choice of negative samples in PCL. mIoU results are reported on the PASCAL-${5^{i}}$ dataset under the 1-shot setting.
	}
	\label{abl_pcl}
	\vspace{-3mm}
	\resizebox{75mm}{!}{
\begin{tabular}{r|ccccc}
\hline
Negative samples & split0 & split1 & split2 & split3 & mean \\ \hline
Non-target Prototypes  & 63.7   & 69.0   & 57.8   & 56.8   & 61.8 \\
BG Prototypes        & 64.4   & 68.7   & 58.4   & 58.6   & 62.6 \\
DO Prototypes        & \textbf{65.4}   & \textbf{72.3}   & \textbf{59.4}   & \textbf{59.8}   & \textbf{64.2}\\
\hline
\end{tabular}}
\vspace{-3mm}
\end{table}
\begin{figure}[]
	\includegraphics[scale= 0.47]{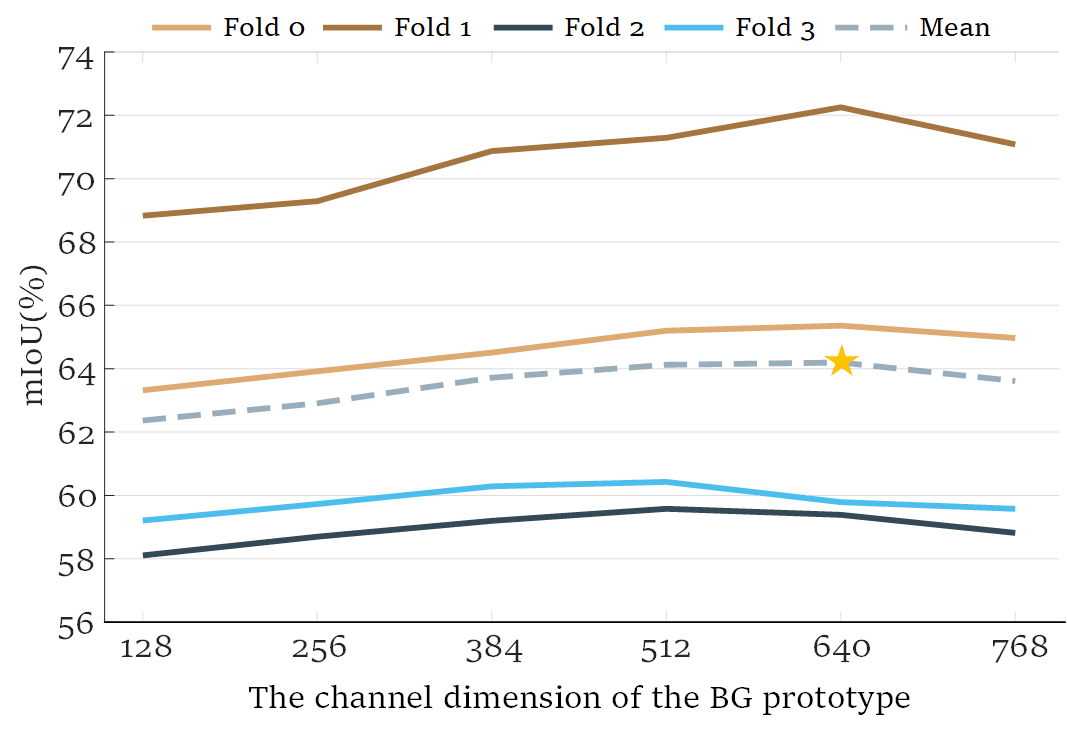}
	\setlength{\abovecaptionskip}{0.cm}
	\caption{Ablation study on the channel dimension of the BG prototype in the 1-shot setting on the PASCAL-${5^{i}}$ dataset. $\bigstar$ indicates the best result of the averaged class mIoU.}
	\label{figureab}
	\vspace{-3mm}
\end{figure}
\vspace{-5mm}

\paragraph{Influence of the Channel Dimension of the BG Prototype.} The channel dimension of the BG prototype is crucial since it determines how much general BG information it can encode. We conduct ablation experiments to explore its optimal value and the results are shown in Figure~\ref{figureab}. It shows that using 512 channels achieves the best performance for fold 2 and 3, while using 640 channels outperforms other values for fold 0, 1, and the mean result. 
Hence, we use 640 channels in the BG prototype in our network setting.
\section{Conclusion}
We address the few-shot semantic segmentation from a new perspective and propose a novel NTRE framework to pay attention to BG and DO regions. We propose the BGMM, BGEM, and DOEM for effectively implementing the mining and eliminating to the BG and DOs. Particularly, the BG mining loss is proposed to supervise the learning of the BGMM and a BG prototype without using BG ground truth. Besides, PCL is proposed to improve the model ability for better distinguishing target objects from DOs. Extensive experiments on two benchmark datasets demonstrate the performance superiority of our method over the previous methods.

\vspace{-2mm}
\paragraph{Acknowledgments:}
This work was supported in part by the Key-Area Research and Development Program of Guangdong Province under Grant 2019B010110001, and the National Natural Science Foundation of China under Grants 62071388, 62136007, U20B2065 and 62036005, the Key R\&D Program of Shaanxi Province under Grant 2021ZDLGY01-08, and the National Key R\&D Program of China under Grant 2020AAA0105701.

{\small
\bibliographystyle{ieee_fullname}
\bibliography{egbib}
}

\end{document}